\documentclass[conference]{IEEEtran}
\IEEEoverridecommandlockouts
\usepackage{cite}
\usepackage{amsmath,amssymb,amsfonts}
\usepackage{graphicx}
\usepackage{textcomp}
\usepackage{xcolor}
\usepackage{pifont}
\usepackage{bbding}
\usepackage{authblk}
\usepackage{xspace}
\usepackage{booktabs}
\usepackage{multirow}
\usepackage{algorithm,algpseudocode}
\usepackage{graphicx}
\usepackage{tikz}
\usetikzlibrary{fit}
\usetikzlibrary{spy}
\usetikzlibrary{arrows.meta}
\usepackage{hyperref}
\def\BibTeX{{\rm B\kern-.05em{\sc i\kern-.025em b}\kern-.08em
    T\kern-.1667em\lower.7ex\hbox{E}\kern-.125emX}}

\makeatletter
\DeclareRobustCommand\onedot{\futurelet\@let@token\@onedot}
\def\@onedot{\ifx\@let@token.\else.\null\fi\xspace}

\makeatother

\begin{document}

\title{HBFormer: A Hybrid-Bridge Transformer for Microtumor and Miniature Organ Segmentation
}

\author[12]{Fuchen Zheng}
\author[2]{Xinyi Chen}
\author[3]{Weixuan Li}
\author[3]{Quanjun Li}
\author[3]{Junhua Zhou}
\author[1]{\authorcr Xiaojiao Guo}
\author[4]{Xuhang Chen}
\author[1*]{Chi-Man Pun\thanks{* Corresponding Authors.}}
\author[2*]{Shoujun Zhou}
\affil[1]{University of Macau}
\affil[2]{Shenzhen Institutes of Advanced Technology, Chinese Academy of Sciences}
\affil[3]{Guangdong University of Technology}
\affil[4]{Huizhou University}

\maketitle

\begin{abstract}
Medical image segmentation is a cornerstone of modern clinical diagnostics. While Vision Transformers that leverage shifted window-based self-attention have established new benchmarks in this field, they are often hampered by a critical limitation: their localized attention mechanism struggles to effectively fuse local details with global context. This deficiency is particularly detrimental to challenging tasks such as the segmentation of microtumors and miniature organs, where both fine-grained boundary definition and broad contextual understanding are paramount. To address this gap, we propose HBFormer, a novel Hybrid-Bridge Transformer architecture. The 'Hybrid' design of HBFormer synergizes a classic U-shaped encoder-decoder framework with a powerful Swin Transformer backbone for robust hierarchical feature extraction. The core innovation lies in its 'Bridge' mechanism, a sophisticated nexus for multi-scale feature integration. This bridge is architecturally embodied by our novel Multi-Scale Feature Fusion (MFF) decoder. Departing from conventional symmetric designs, the MFF decoder is engineered to fuse multi-scale features from the encoder with global contextual information. It achieves this through a synergistic combination of channel and spatial attention modules, which are constructed from a series of dilated and depth-wise convolutions. These components work in concert to create a powerful feature bridge that explicitly captures long-range dependencies and refines object boundaries with exceptional precision. Comprehensive experiments on challenging medical image segmentation datasets, including multi-organ, liver tumor, and bladder tumor benchmarks, demonstrate that HBFormer achieves state-of-the-art results, showcasing its outstanding capabilities in microtumor and miniature organ segmentation. Code and models are available at: \url{https://github.com/lzeeorno/HBFormer}. 
\end{abstract}

\begin{IEEEkeywords}
Medical Image Segmentation, Tumor Segmentation, Vision Transformer, Attention Mechanism, Multi-scale Feature Fusion
\end{IEEEkeywords}

\section{Introduction}
Medical image segmentation, a critical task in computer-assisted diagnosis, demands the precise delineation of anatomical structures. A fundamental challenge, especially for small targets like microtumors, is the need for models to perceive both fine-grained local details for boundary definition and high-level global context for robust recognition.

While Convolutional Neural Networks (CNNs) like the U-Net~\cite{4unet} are foundational for capturing local features, their inherent locality limits their ability to model long-range spatial dependencies. Vision Transformers (ViTs) were adapted to overcome this, with models like Swin-Unet~\cite{21swinUnet2022} demonstrating the power of combining a U-shaped architecture with a Swin Transformer~\cite{37swin} backbone. However, the very efficiency of Swin Transformer, derived from its shifted window-based self-attention, reintroduces a form of locality. This can impede the seamless integration of features across different scales, creating a new bottleneck for tasks demanding a holistic image understanding.

To bridge this crucial gap between local feature extraction and global context aggregation in window-based models, we propose HBFormer. Our architecture is a hybrid CNN-Transformer design that leverages a Swin Transformer encoder within a U-shaped framework. The central contribution is its innovative Multi-Scale Feature Fusion (MFF) decoder. This decoder acts as an explicit bridge, meticulously designed to fuse multi-scale feature maps from the encoder with rich, long-range contextual information, compensating for the localized scope of the encoder's windowed attention.

Our main contributions are summarized as follows:
\begin{enumerate}
\item We propose a novel hybrid architecture, \textbf{HBFormer}, that effectively integrates a Swin Transformer encoder with an innovative ``bridge'' decoder to resolve the conflict between local attention efficiency and global context necessity in medical image segmentation.
\item We design a \textbf{Multi-Scale Feature Fusion (MFF) decoder} that acts as this bridge, employing a unique combination of attention mechanisms to explicitly capture long-range dependencies and fuse them with fine-grained features for precise boundary delineation.
\item We introduce the \textbf{Medical-Deformable Spatial Pyramid Pooling (Med-DSPP) Block}. This block optimizes the dilation rate ratios to effectively restore the intricate and often distorted shapes of organs and tumors. The Med-DSPP Block integrating the adaptive sampling grid of deformable convolution with a parallel multi-rate dilated convolution allows the network to dynamically adjust its receptive field to preserve and recover complex geometric features and expands the contextual view to effectively cover all organs and tumors located at the margins.

\end{enumerate}


\begin{figure}[ht]
\centerline{\includegraphics[width=\linewidth]{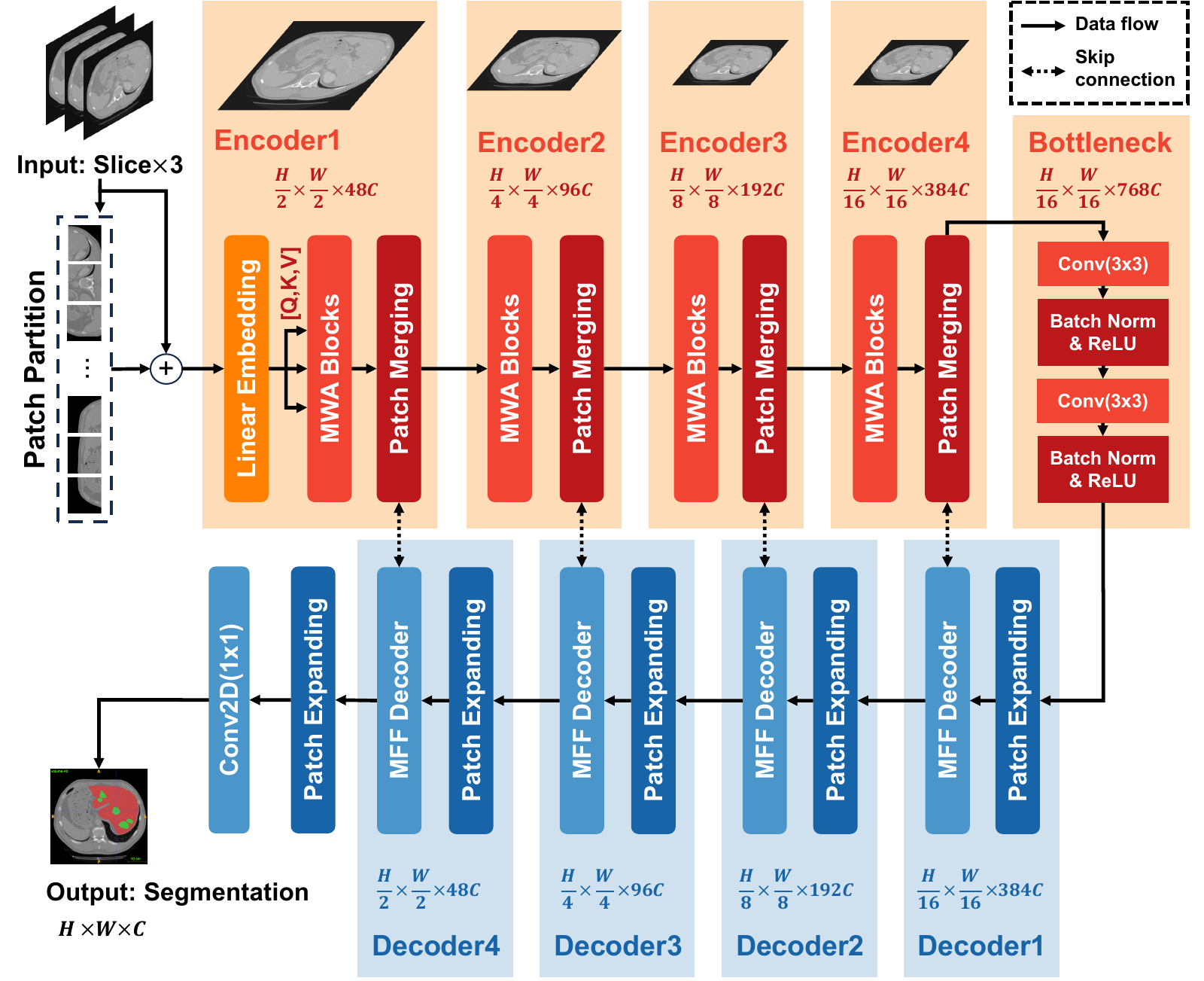}}
\caption{Overview of the HBFormer architecture.}
\label{fig1}
\end{figure}

\section{Related Work}
This section reviews the pivotal advancements, tracing the trajectory from convolutional foundations to the latest hybrid Transformer designs, and highlighting the specific challenges that motivate our work.

\subsection{Convolutional Backbones and the U-Net Family}
The U-Net architecture~\cite{4unet} established a dominant paradigm in medical image segmentation. Its symmetric encoder-decoder structure, augmented with skip connections, proved exceptionally effective at capturing fine-grained details while preserving spatial context. This foundational design inspired a plethora of variants. For instance, ResUNet~\cite{6resunet} incorporated residual connections~\cite{7resnet} to facilitate the training of deeper networks and improve information flow, while UNet++~\cite{13unet++2018} introduced nested and dense skip connections to bridge the semantic gap between the encoder and decoder. Despite their success, these CNN-based models are fundamentally constrained by the locality of the convolution operation, which limits their ability to explicitly model long-range dependencies—a critical requirement for understanding global context in complex scenes.

\subsection{Vision Transformers and Hybrid Architectures}
To address the locality constraint of CNNs, the Vision Transformer (ViT)~\cite{34vit} adapted the Transformer architecture for computer vision, demonstrating its power in capturing global relationships via self-attention mechanisms. Recognizing the complementary nature of CNNs (strong local feature extraction) and Transformers (strong global context modeling), a wave of hybrid models emerged. For instance, the very mechanism that grants Swin-Unet~\cite{21swinUnet2022} its efficiency—the windowed attention—reintroduces a form of locality. The information exchange between different windows and scales is implicit and constrained, potentially creating a bottleneck for the seamless fusion of multi-scale features. This is particularly problematic for segmenting small or intricate structures where integrating fine-grained boundary details with a holistic contextual understanding is paramount.

Therefore, a critical challenge remains: how to design an architecture that retains the computational efficiency of windowed attention while explicitly and effectively bridging the gap between multi-scale local features and global contextual information. Our work, HBFormer, is designed to directly address this challenge.

\section{Methodology}
This section details the architecture of HBFormer, our proposed Hybrid-Bridge Transformer. We first present the overall U-shaped architecture that integrates a powerful encoder and our novel decoder. We then elaborate on the two core components: (1) the Multi-scale Window Attention (MWA) Transformer block, the fundamental building block of our Swin Transformer-based encoder, and (2) the Multi-Scale Feature Fusion (MFF) decoder, which acts as a sophisticated bridge to synergize features across multiple scales for precise segmentation.

\subsection{HBFormer Architecture}
As illustrated in Figure~\ref{fig1}, HBFormer adopts a hierarchical U-shaped architecture. This design leverages an encoder to capture rich, hierarchical features and a decoder to progressively reconstruct high-resolution segmentation masks, with skip connections bridging corresponding levels to preserve fine-grained details.
\subsection{Encoder: MWA Transformer Block}
\begin{figure}[ht]
\centerline{\includegraphics[width=0.8\columnwidth]{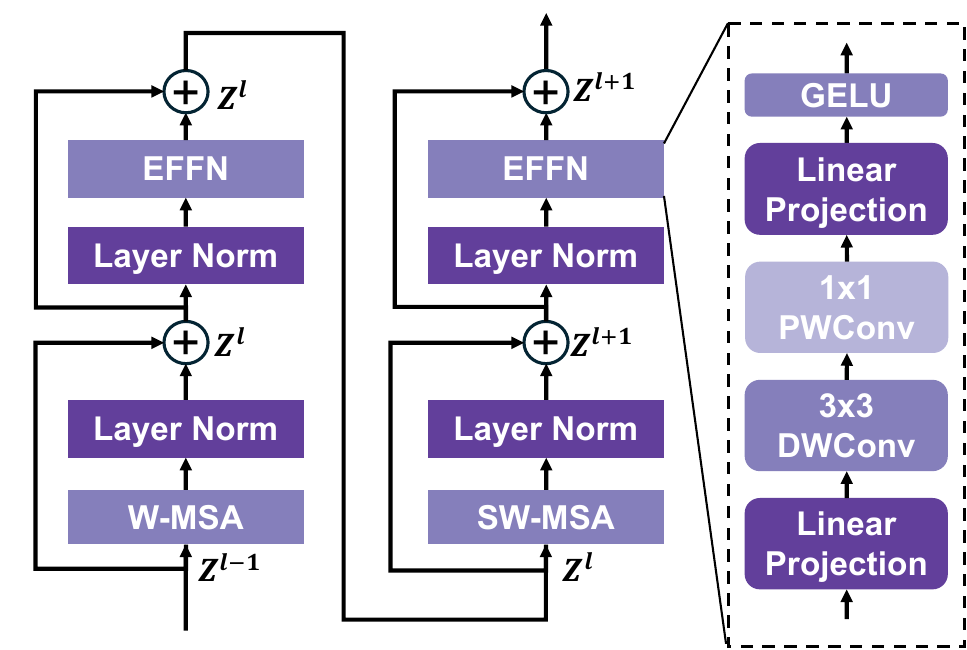}}
\caption{This figure presents details of a schematic diagram of the proposed Multi-scale Window Attention (MWA) transformer block.}
\label{fig2}
\end{figure}

The core component of our encoder is the Multi-scale Window Attention (MWA) Transformer block, which is an enhanced version of the standard Swin Transformer block. As shown in Figure~\ref{fig2}, a consecutive pair of MWA blocks replaces the standard multi-head self-attention (MSA) with window-based (W-MSA) and shifted-window-based (SW-MSA) attention, respectively. This retains the computational efficiency of the Swin Transformer.

Our key innovation within the encoder lies in the Enhanced Feed-Forward Network (EFFN), which replaces the standard Feed-Forward Network (FFN). Recognizing that standard MLPs in FFNs are limited in capturing local context~\cite{70MLP2021cvt, 71eighboringpixels2021CVPR}, we explicitly inject convolutional inductive biases into the Transformer block. As depicted in Figure~\ref{fig2}, the EFFN first projects the input tokens to a higher dimension with a linear layer. These are then reshaped into 2D feature maps and processed by a sequence of a $3\times3$ depth-wise convolution and a $1\times1$ point-wise convolution. This operation efficiently captures local spatial context. The features are then reshaped back into tokens and projected to their original dimension. This injects local inductive bias while preserving the Transformer’s global modeling capacity.

The computation within two consecutive MWA Transformer blocks is formulated as:
\begin{equation}
\begin{aligned}
& \hat{X}^{l} = \operatorname{W-MSA}(\operatorname{LN}(X^{l-1})) + X^{l-1}, \\
& X^{l} = \operatorname{EFFN}(\operatorname{LN}(\hat{X}^{l})) + \hat{X}^{l},  \\
& \hat{X}^{l+1} = \operatorname{SW-MSA}(\operatorname{LN}(X^{l})) + X^{l}, \\
& X^{l+1} = \operatorname{EFFN}(\operatorname{LN}(\hat{X}^{l+1})) + \hat{X}^{l+1}, 
\end{aligned}
\end{equation}
where W-MSA and SW-MSA denote windowed and shifted-windowed multi-head self-attention, respectively. We also incorporate a relative position bias $B \in \mathbb{R}^{(2M-1) \times (2M-1)}$ into the self-attention computation, which has proven beneficial in prior works~\cite{12swin2021}:
\begin{equation}
\operatorname{Attention}(Q, K, V) = \operatorname{SoftMax}\left(\frac{QK^T}{\sqrt{d}} + B\right)V,
\end{equation}
where $Q, K, V$ are the query, key, and value matrices, and $d$ is the query/key dimension.

\subsection{The MFF Bridge: A Multi-Scale Feature Fusion Decoder}
\begin{figure}[ht]
\centerline{\includegraphics[width=0.9\columnwidth]{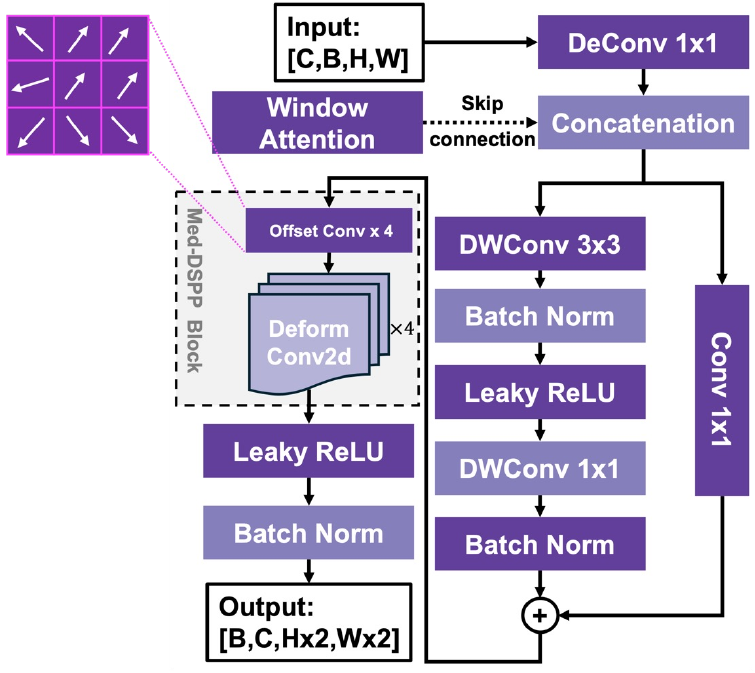}}
\caption{This figure presents details of a schematic diagram of the proposed Multi-scale Feature Fusion (MFF) Decoder.}
\label{fig3}
\end{figure}
A significant limitation of window-based Transformers lies in their decoders, which typically rely on simplistic upsampling and convolution layers. These components are inadequate for effectively fusing multi-scale features and compensating for the locality inherent in the encoder's attention mechanism~\cite{12swin2021,49swinUNETR}. To overcome this limitation, we introduce the Multi-Scale Feature Fusion (MFF) decoder, designed as an explicit bridge to robustly integrate features across varying scales.

As illustrated in Figure~\ref{fig3}, the MFF module begins by upsampling the input from the preceding decoder layer and concatenating it with skip-connection features from the corresponding encoder stage. This fused feature map is subsequently processed through a series of Depth-wise Convolutions and the Med-DSPP block, allowing sophisticated extraction and combination of information.

Inspired by the success of ASPP~\cite{41ASPP} in capturing multi-scale context and Deformable Convolution~\cite{84deformConV2017deformable} in modeling geometric transformations, we designed one Medical Deformable Spatial Pyramid Pooling (Med-DSPP) Block. The Med-DSPP contains four parallel branches. Each branch consist by a 3×3 deformable convolution (offset kernel 3×3, offsets initialized to zero) + 3×3 dilated convolutions with rates r $\in${1, 6, 12, 18}, followed by BN + LeakyReLU. Outputs are concatenated and fused via a 1×1 Conv, with a residual connection added from the input of Med-DSPP. Channel widths per decoder stage are {C1, C2, C3, C4} = {96, 192, 384, 768}, aligned to the encoder. This design not only allows the network to dynamically adjust its receptive field to preserve and restore complex geometric features but also expands the contextual view to effectively cover organs and tumors situated in peripheral locations.

\section{Experiments}
In this section, we present a comprehensive experimental evaluation designed to rigorously validate the effectiveness and superiority of our proposed HBFormer. We first outline the datasets, evaluation metrics, and implementation details. Subsequently, we benchmark HBFormer against a range of state-of-the-art methods in medical image segmentation. Finally, we conduct detailed ablation studies to scrutinize the contribution of each core component within the HBFormer architecture.

\begin{table*}[ht]
\centering
\caption{Comparison with State-of-the-Art models on the ISICDM2019 and LITS2017 datasets. The best results are bolded while the second best are underlined. \textbf{Statistical notes:} HBFormer results are reported as Mean $\pm$ Std across 3 seeds. Cross-paper comparisons are descriptive due to dimensionality and pipeline differences. We report mean $\pm$ std over 3 seeds for Significance tests.}
\resizebox{0.9\textwidth}{!}{%
\begin{tabular}{c|cc|cc|cc|cc}
\toprule
\multirow{3}{*}{Method} & \multicolumn{4}{c|}{ISICDM2019}                                    & \multicolumn{4}{c}{LITS2017}                                      \\ \cline{2-9} 
                         & \multicolumn{2}{c|}{Average}     & Bladder        & Tumor          & \multicolumn{2}{c|}{Average}     & Liver        & Tumor          \\ 
                          & DSC(\%) $\uparrow$        & mIoU(\%) $\uparrow$       & DSC(\%) $\uparrow$        & DSC(\%) $\uparrow$        & DSC(\%) $\uparrow$        & mIoU(\%) $\uparrow$       & DSC(\%) $\uparrow$        & DSC(\%) $\uparrow$        \\ 
                         \midrule
                         \midrule
ViT~\cite{34vit}+CUP~\cite{15chen2021transunet}                 & 88.60 $\pm$ 0.18          & 84.40 $\pm$ 0.18         & 91.88 $\pm$ 0.15          & 85.32 $\pm$ 0.20          & 80.33 $\pm$ 0.20         & 77.25 $\pm$ 0.22          & 83.97 $\pm$ 0.18         & 76.69  $\pm$ 0.22        \\

R50-ViT~\cite{34vit}+CUP~\cite{15chen2021transunet}             & 88.77 $\pm$ 0.18         & 85.62 $\pm$ 0.18         & 92.05 $\pm$ 0.15         & 85.49 $\pm$ 0.20         & 82.62 $\pm$ 0.20         & 79.68 $\pm$ 0.21         & 85.83 $\pm$ 0.18         & 79.41 $\pm$ 0.22         \\
ResUNet++~\cite{53resunet++}           & 87.11 $\pm$ 0.20         & 83.78 $\pm$ 0.20         & 89.90 $\pm$ 0.18         & 84.32 $\pm$ 0.22         & 75.73 $\pm$ 0.22 & 74.19 $\pm$ 0.22 & 79.12 $\pm$ 0.20 & 72.34 $\pm$ 0.24             \\
ResT-V2-B~\cite{resT2022v2}            & 89.26 $\pm$ 0.19         & 82.13 $\pm$ 0.22         & 93.01 $\pm$ 0.16         & 85.50 $\pm$ 0.21         & 78.53 $\pm$ 0.21 & 75.24 $\pm$ 0.21 & 81.22 $\pm$ 0.19 & 75.83 $\pm$ 0.23           \\
TransUNet~\cite{15chen2021transunet}
& 92.57 $\pm$ 0.16 & 89.56 $\pm$ 0.17 & 95.28 $\pm$ 0.14 & 89.86 $\pm$ 0.18 & 91.47 $\pm$ 0.16 & 88.97 $\pm$ 0.15  & 93.61 $\pm$ 0.14 & \underline{89.33 $\pm$ 0.18}               \\
SwinUNet~\cite{21swinUnet2022}
& 91.95 $\pm$ 0.17 & 89.77 $\pm$ 0.17 & 94.73 $\pm$ 0.15 & 89.17 $\pm$ 0.18 & 89.68 $\pm$ 0.17 & 86.62 $\pm$ 0.17 & 93.31 $\pm$ 0.15 & 86.04 $\pm$ 0.19               \\
Swin UNETR~\cite{49swinUNETR}
& 92.60 $\pm$ 0.15 & \underline{90.61 $\pm$ 0.15} & 95.08 $\pm$ 0.14 & 90.12 $\pm$ 0.16 & \underline{91.95 $\pm$ 0.16} & \underline{90.02 $\pm$ 0.14} & \underline{94.73 $\pm$ 0.13} & 85.76 $\pm$ 0.18               \\
UNETR~\cite{48unetr}                & 91.55 $\pm$ 0.18         & 88.34 $\pm$ 0.17         & 94.83 $\pm$ 0.15         & 88.26 $\pm$ 0.18         & 89.38 $\pm$ 0.17 & 87.46 $\pm$ 0.17 & 92.89 $\pm$ 0.15 & 85.86 $\pm$ 0.19               \\
nnFormer~\cite{47nnformer}               & \underline{93.54 $\pm$ 0.14}         & 89.11 $\pm$ 0.17         & \underline{96.97 $\pm$ 0.12}         & \underline{90.41 $\pm$ 0.16}         & 91.74 $\pm$ 0.15 & 89.95 $\pm$ 0.15 & 94.57 $\pm$ 0.13 & 88.91 $\pm$ 0.17              \\
\textbf{HBFormer (Ours)} & \textbf{96.33 $\pm$ 0.10} & \textbf{94.74 $\pm$ 0.10} & \textbf{98.97 $\pm$ 0.06} & \textbf{93.69 $\pm$ 0.14} & \textbf{94.82 $\pm$ 0.12} & \textbf{91.98 $\pm$ 0.10} & \textbf{95.89 $\pm$ 0.08} & \textbf{93.74 $\pm$ 0.16} \\
\bottomrule
\end{tabular}
}
\label{tab:comp}
\end{table*}

\begin{table*}[ht]
    \centering
    \caption{Comparison with State-of-the-Art models on the Synapse multi-organ dataset. The best results are bolded while the second best are underlined. We report mean $\pm$ std over 3 seeds for Significance tests. \textbf{Statistical notes:} HBFormer results are reported as Mean $\pm$ Std across 3 seeds. Cross-paper comparisons are descriptive due to dimensionality and pipeline differences.
    }
    \resizebox{0.9\textwidth}{!}{%
    \begin{tabular}{c|c|c|c|c|c|c|c|c|c}
        \toprule
        Model  & \multicolumn{1}{c|}{Average} & Aotra & Gallbladder & Kidney(Left) & Kidney(Right) & Liver & Pancreas & Spleen & Stomach \\
               & DSC(\%)$\uparrow$ & DSC(\%)$\uparrow$ & DSC(\%)$\uparrow$ & DSC(\%)$\uparrow$ & DSC(\%)$\uparrow$ & DSC(\%)$\uparrow$ & DSC(\%)$\uparrow$ & DSC(\%)$\uparrow$ & DSC(\%)$\uparrow$\\
        \midrule
        \midrule

        ViT~\cite{34vit}+CUP~\cite{15chen2021transunet}    
            & 67.86 $\pm$ 0.21 & 70.19 $\pm$ 0.20  & 45.10 $\pm$ 0.24  & 74.70 $\pm$ 0.20  & 67.40 $\pm$ 0.22  & 91.32 $\pm$ 0.15  & 42.00 $\pm$ 0.25  & 81.75 $\pm$ 0.18  & 70.44 $\pm$ 0.21      \\
    
        R50-ViT~\cite{34vit}+CUP~\cite{15chen2021transunet}  
            & 71.29 $\pm$ 0.19  & 73.73 $\pm$ 0.19 & 55.13 $\pm$ 0.24 & 75.80 $\pm$ 0.20 & 72.20 $\pm$ 0.21 & 91.51 $\pm$ 0.15 & 45.99 $\pm$ 0.18 & 81.99 $\pm$ 0.13 & 73.95 $\pm$ 0.19     \\
            
        TransUNet~\cite{15chen2021transunet}
            & 83.85 $\pm$ 0.17 & 90.68 $\pm$ 0.21 & \textbf{71.99} $\pm$ 0.23 & \underline{86.04 $\pm$ 0.19}  & 83.71 $\pm$ 0.18 & 95.54 $\pm$ 0.11 & 73.96 $\pm$ 0.18 & \textbf{88.80 $\pm$ 0.15}  & 80.06 $\pm$ 0.13      \\
            
        SwinUNet~\cite{21swinUnet2022}
            & 78.44 $\pm$ 0.19 & 85.47 $\pm$ 0.21 & 66.53 $\pm$ 0.23 & 83.28 $\pm$ 0.20 & 79.61 $\pm$ 0.19 & 94.29 $\pm$ 0.12 & 56.58 $\pm$ 0.21 & 85.16 $\pm$ 0.14 & 76.60 $\pm$ 0.18    \\

        UNETR~\cite{48unetr}
            & 79.57 $\pm$ 0.18 & 89.99 $\pm$ 0.18 & 60.56 $\pm$ 0.24 & 85.66 $\pm$ 0.18 & 84.80 $\pm$ 0.17 & 94.46 $\pm$ 0.13 & 59.25 $\pm$ 0.20 & \underline{87.81 $\pm$ 0.13} & 73.99 $\pm$ 0.18     \\
        

        CoTr~\cite{52cotr}
            & 85.26 $\pm$ 0.17 & \textbf{92.96 $\pm$ 0.21} & 71.09 $\pm$ 0.22 & 85.70 $\pm$ 0.19 & 85.71 $\pm$ 0.19 & \underline{96.88 $\pm$ 0.11} & 81.28 $\pm$ 0.15 & 86.74 $\pm$ 0.13 & \underline{81.74 $\pm$ 0.13}     \\

        nnFormer~\cite{47nnformer}
            & \underline{85.30 $\pm$ 0.16} & 90.72 $\pm$ 0.19 & 71.67 $\pm$ 0.23 & 85.60 $\pm$ 0.17 & \underline{87.02 $\pm$ 0.17} &  96.28 $\pm$ 0.12 & \textbf{82.12 $\pm$ 0.15} &  87.30 $\pm$ 0.13 & 81.69 $\pm$ 0.13     \\
        \vspace{0.2em}
        \textbf{HBFormer (Ours)}
            & \textbf{86.09 $\pm$ 0.16} & \underline{92.13 $\pm$ 0.18} & \underline{71.80 $\pm$ 0.22} & \textbf{87.27 $\pm$ 0.17} & \textbf{89.11 $\pm$ 0.16} & \textbf{97.01 $\pm$ 0.11} & \underline{81.87 $\pm$ 0.15} & 87.71 $\pm$ 0.13 & \textbf{81.82 $\pm$ 0.13}     \\
    \bottomrule
    \end{tabular}
    }
    \label{tab:syna}
\end{table*}

\subsection{Datasets and Implementation Details}\label{sec:4a}
To ensure a thorough and fair evaluation, we conducted experiments on three publicly available and widely recognized medical image segmentation datasets:
\begin{enumerate}
\item \textbf{LiTS2017}~\cite{44lits2017}: A challenging dataset for liver and liver tumor segmentation, comprising 131 contrast-enhanced 3D abdominal CT scans.
\item \textbf{ISICDM2019}~\cite{46ISICDM2019}: A dataset focused on bladder tumor segmentation, containing 2200 bladder and tumor images.
\item \textbf{Synapse}~\cite{45synapse}: This dataset comprises 30 cases of abdominal CT scans. Following the partitioning method employed in~\cite{47nnformer}, 18 cases are selected to construct the training set, while the remaining 12 cases are utilized for testing.
\end{enumerate}

To ensure robust and unbiased evaluation, we employed a five-fold cross-validation methodology for ISICDM2019 and LiTS2017. For Synapse we follow the standardized splits adopted by Swin-UNet~\cite{21swinUnet2022} to ensure comparability (no additional cross-validation). All CT slices or 2.5D stacks are resized to $512\times512$.

Our implementation of HBFormer utilizes the PyTorch framework, with all models trained on a single NVIDIA GeForce RTX 4090 GPU. We employed the Stochastic Gradient Descent (SGD) optimizer~\cite{SGD2010large} with a momentum of 0.98 and a weight decay of $1 \times 10^{-6}$. The learning rate was initialized to $1 \times 10^{-2}$ and managed by a cosine decay schedule down to a minimum of $6 \times 10^{-6}$. Standard data augmentation techniques, including random horizontal flipping and rotation, were applied during training. The encoder of our HBFormer was initialized with weights pre-trained on Swin Transformer~\cite{37swin}, a common practice to accelerate convergence and improve performance. Statistical analysis uses three random seeds in every experiment. We train each configuration with 3 independent seeds (3407, 8261, 10993).

\subsection{Objective Function}
HBFormer is trained with the BCE–Dice loss $\mathcal{L}_{BD}$~\cite{62BCEDiceLoss2016v}, a composite loss that simply sums the Binary Cross-Entropy (BCE) loss $\mathcal{L}_{BCE}$ and the Dice loss $\mathcal{L}_{D}$. The BCE term drives pixel-wise classification, whereas the Dice term directly maximises the spatial overlap between prediction and ground truth, enabling the network to learn both fine-grained accuracy and shape consistency. 

\subsection{Evaluation Metrics}
We assess segmentation performance using two standard and widely adopted metrics:

\textbf{Dice Similarity Coefficient (DSC)}~\cite{64dice1945measures}: DSC quantifies the overlap between the predicted segmentation ($P$) and the ground truth ($G$). It is a primary metric for segmentation accuracy, with higher values indicating better performance.

\textbf{Mean Intersection over Union (mIoU)~\cite{76IoU2010pascal}}: mIoU computes the average ratio of the intersection to the union of the predicted and ground-truth regions across all classes ($C$). It provides a stringent measure of segmentation quality.

\begin{figure*}[ht]
\centerline{\includegraphics[width=0.9\linewidth]{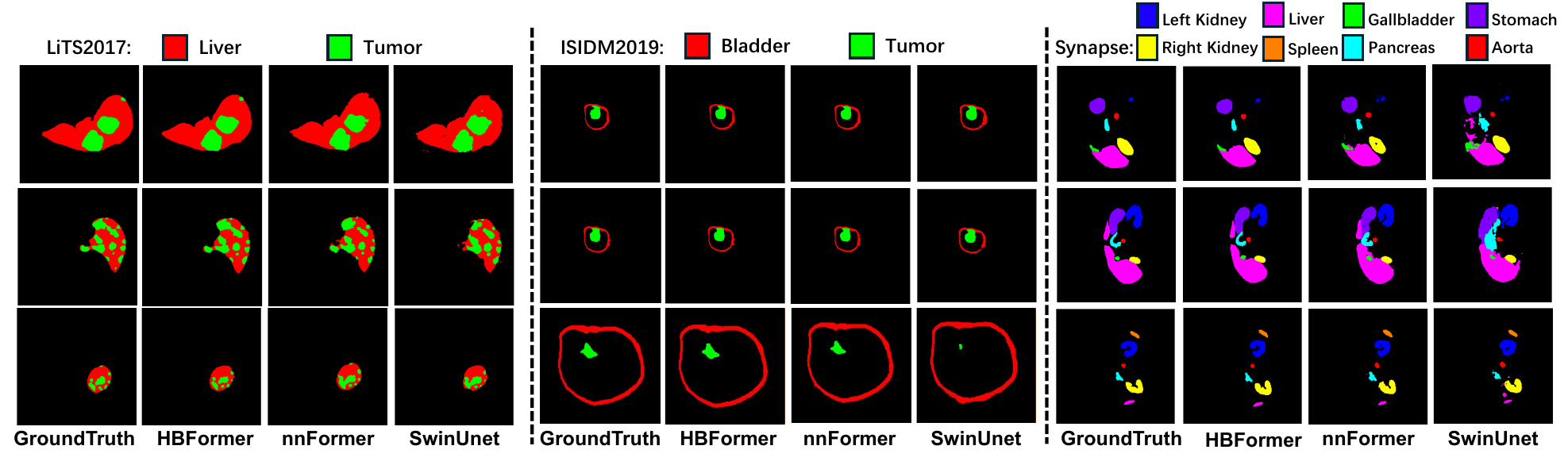}}
\caption{LiTS2017, ISICDM2019 and Synapse Prediction Results.}
\label{fig4}
\end{figure*}

\subsection{Comparisons with State-of-the-Art Methods}
This subsection details the comparative performance of HBFormer against leading methods on the LiTS2017, ISICDM2019, and Synapse datasets. The quantitative results are summarized in Table~\ref{tab:comp} and~\ref{tab:syna}.

\subsubsection{Liver Tumor Segmentation}
Liver Tumor Segmentation on LiTS2017
As demonstrated in Table~\ref{tab:comp}, HBFormer establishes new state-of-the-art performance on the LiTS2017 dataset, achieving an average DSC of 94.82\% and mIoU of 91.98\%. Notably, our method attains 93.74\% DSC for tumor segmentation. These results represent substantial improvements over the previous best-performing method, Swin UNETR, which achieved 91.95\% average DSC and 90.02\% mIoU. The 2.87\% improvement in average DSC and 1.96\% enhancement in mIoU demonstrate the effectiveness of our Multi-Scale Feature Fusion (MFF) bridge decoder in preserving critical spatial information across different scales.

\subsubsection{Bladder Tumor Segmentation}
On the ISICDM2019 bladder tumor dataset, HBFormer demonstrates exceptional performance with an average DSC of 96.33\% and mIoU of 94.74\%(Table~\ref{tab:comp}). The method achieves 98.97\% DSC for bladder segmentation and 93.69\% DSC for tumor segmentation. Compared to the second-best performing nnFormer (93.54\% average DSC), our approach delivers a notable 2.79\% improvement in overall performance.

The most clinically relevant advancement is observed in tumor segmentation accuracy, where HBFormer achieves 93.69\% DSC compared to nnFormer's 90.41\%, representing a 3.28\% improvement. This enhancement is attributed to the synergistic integration of our Multi-Window Attention (MWA) mechanism and the MFF bridge decoder, which collectively enable more precise capture of complex tumor boundaries and subtle morphological variations characteristic of bladder malignancies.

\subsubsection{Multi-Organ Segmentation}
On the highly challenging Synapse multi-organ dataset (Table~\ref{tab:syna}), HBFormer continues to demonstrate its power and robustness. HBFormer achieves the highest overall average DSC of 86.09\%, surpassing the previous best method nnFormer (85.30\%) by 0.79\%. This improvement demonstrates consistent performance across anatomically diverse structures with varying scales and morphological characteristics.

In conclusion, HBFormer establishes new benchmarks across three challenging medical segmentation tasks through its innovative architectural design. The clinical significance of these improvements cannot be understated. In medical image segmentation, even marginal improvements in accuracy translate to enhanced diagnostic confidence, reduced inter-observer variability, and ultimately improved patient outcomes. HBFormer's consistent performance across diverse anatomical structures and pathological conditions positions it as a robust solution for clinical deployment in computer-aided diagnosis systems.

\subsection{Visualization of Segmentation Results}
To provide a qualitative assessment of HBFormer's capabilities, we present visual comparisons against other leading methods on all three datasets.

Figure~\ref{fig4} showcases segmentation results on the LiTS2017, ISICDM2019 and Synapse multi-organ dataset. The qualitative superiority of HBFormer is evident. In the figures, competing models often fail to detect small tumor nodules (the green areas disappear), a consequence of information loss in deep network layers. In contrast, HBFormer, powered by its MFF bridge, successfully preserves these critical features, leading to a more complete and accurate segmentation that is closer to the ground truth. In addition, HBFormer produces consistently accurate segmentations across diverse organs, including the pancreas—an organ notoriously difficult to segment. 


\begin{center}
\begin{table}[ht]
    \centering
    \caption{Ablation study of different modules in HBFormer. \textbf{Statistical notes:} HBFormer results are reported as Mean $\pm$ Std across 3 seeds.}
    \resizebox{\columnwidth}{!}{
    \begin{tabular}{c c c c|c|c}
        \toprule
        SwinUnet~\cite{21swinUnet2022} & MFF  & FFN  & EFFN &
        ISICDM2019 & 
        LiTS2017 \\
        Decoder & Decoder & Encoder & Encoder &  Average DSC $\uparrow$ & Average DSC $\uparrow$ \\
        
        \midrule
    
            \checkmark & $\times$ &  \checkmark & $\times$ & 88.95\% $\pm$ 0.18 & 87.27\% $\pm$ 0.15\\
        
            $\times$ & \checkmark & \checkmark & $\times$ & 
            90.80\% $\pm$ 0.11 & 89.14\%  $\pm$ 0.13\\
        
             \checkmark & $\times$ & $\times$ & \checkmark & 91.32\%  $\pm$ 0.10 & 90.37\%  $\pm$ 0.12\\

            $\times$ & \checkmark & $\times$ & \checkmark & 94.19\%  $\pm$ 0.19 & 92.47\%  $\pm$ 0.18\\

        \bottomrule
    \end{tabular}
    }
    \label{tab:ablation}
\end{table}
\end{center}

\subsection{Ablation Study}
To systematically evaluate the contribution of each architectural component in HBFormer, we conducted a comprehensive ablation study on the ISICDM2019 and LiTS2017 datasets. The experimental design isolates the effects of our two key innovations: the Enhanced Feed-Forward Network (EFFN) in the encoder and the Multi-Scale Feature Fusion (MFF) bridge decoder. Table~\ref{tab:ablation} presents the quantitative results of four distinct architectural configurations.
We begin our analysis with a baseline configuration employing the SwinUNet decoder combined with a standard FFN encoder. This configuration achieves 88.95\% DSC on ISICDM2019 and 87.27\% DSC on LiTS2017, establishing our performance baseline for subsequent component evaluations.
Replacing the SwinUNet decoder with our MFF decoder while keeping the standard FFN encoder yields 90.80\% (+1.85) on ISICDM2019 and 89.14\% (+1.87) on LiTS2017 over the baseline (88.95\% / 87.27\%). Introducing only EFFN (row 3) improves to 91.32\% and 90.37\% (+2.37 / +3.10). Combining both MFF and EFFN (row 4) further boosts performance to 94.19\% and 92.47\%, absolute gains of +5.24\% and +5.20\%. Adding our Multi-Window Attention (not shown in Table~\ref{tab:ablation}) further improves the scores to 96.33\% (ISICDM2019) and 94.82\% (LiTS2017) as shown in Table~\ref{tab:comp}. Significance testing is restricted to models re-implemented under our pipeline; detailed per-seed paired tests for ablations are reported in the supplement.

The ablation results reveal that while both components contribute significantly to performance enhancement, their combination yields greater improvements than the sum of their individual contributions.

\section{Conclusion}
We presented HBFormer, a hybrid CNN–Transformer for high-precision medical image segmentation. An EFFN injects convolutional inductive bias into Transformer blocks, while an MFF bridge decoder fuses multi-scale features to preserve fine details. Across LiTS2017, ISICDM2019, and Synapse, HBFormer consistently surpasses state-of-the-art methods, with ablations showing cumulative Dice gains of +5.24/+5.20 from MFF+EFFN and up to +7.38/+7.55 with the full model (96.33\% vs 88.95\%; 94.82\% vs 87.27\%). HBFormer effectively balances global context and local precision, indicating strong potential for clinical deployment. Future work will extend to 3D imaging and model compression for resource-constrained settings.

\section{Acknowledgment}
This work was supported in part by Shenzhen Medical Research Fund (No. D2404001), in part by the Science and Technology Development Fund, Macau SAR, under Grant 0193/2023/RIA3 and 0079/2025/AFJ, and the University of Macau under Grant MYRG-GRG2024-00065-FST-UMDF, and Key-Area Research and Development Program of Guangdong Province(No. 2025B1111020001), in part by the Natural Science Foundation of Guangdong Province (No. 2023A1515010673), in part by the Shenzhen Science and Technology Innovation Bureau key project (No. JSGG20220831110400001, No. CJGJZD20230724093303007,KJZD20240903101259001), in part by Shenzhen Engineering Laboratory for Diagnosis \& Treatment Key Technologies of Interventional Surgical Robots (XMHT20220104009), and the Key Laboratory of Biomedical Imaging Science and System, CAS, for the Research platform support.

\bibliographystyle{IEEEtran}
\bibliography{ref}

\end{document}